\documentclass[sigconf,natbib=false, balance=false]{acmart}

\AtBeginDocument{%
  }

\acmConference[arXiv]{Preprint}{2026}{}

\setcopyright{None}

\RequirePackage[
  datamodel=acmdatamodel,
  style=acmnumeric,
]{biblatex}

\usepackage{multirow}
\usepackage{multicol}
\usepackage{amsmath}

\usepackage{amssymb}
\usepackage{algorithm}
\usepackage{algorithmic}
\usepackage{tikz}
\usetikzlibrary{positioning, fit, backgrounds, shapes.geometric, arrows.meta, calc}
\usepackage{pgfplots}
\pgfplotsset{compat=1.18}
\usepackage{booktabs}
\usepackage{xcolor}
\usepackage{xspace}
\usepackage{listings}
\usepackage{soul}

\usepackage[disable]{todonotes}

\newcommand{\eg}{e.g.\xspace}

\lstdefinestyle{codestyle}{
  basicstyle=\ttfamily\footnotesize,
  breaklines=true,
  frame=single,
  numbers=left,
  numberstyle=\tiny\color{gray},
  xleftmargin=2em,
  framexleftmargin=1.5em,
  keywordstyle=\color{blue!70!black},
  commentstyle=\color{green!50!black},
  stringstyle=\color{red!60!black},
  showstringspaces=false,
  tabsize=4,
}

\lstdefinestyle{promptstyle}{
  basicstyle=\ttfamily\scriptsize,
  breaklines=true,
  frame=single,
  numbers=none,
  xleftmargin=1em,
  framexleftmargin=0.5em,
  keywordstyle={},
  commentstyle={},
  stringstyle={},
  showstringspaces=false,
  tabsize=2,
  columns=fullflexible,
}

\begin{document}
\title{Selective Left-Shift: Turning Test-Time Compute and Difficulty-based Curation into Training Data for Low-Resource Code Generation}

\author{Didula Samaraweera}
\affiliation{%
  \institution{WSO2}
  \city{Santa Clara}
  \state{CA}
  \country{USA}
}
\email{didula@wso2.com}
\orcid{0009-0000-7712-6056}

\author{Anjana Supun}
\affiliation{%
  \institution{WSO2}
  \city{Santa Clara}
  \state{CA}
  \country{USA}
}
\email{anjanas@wso2.com}

\author{Srinath Perera}
\affiliation{%
  \institution{WSO2}
  \city{Santa Clara}
  \state{CA}
  \country{USA}}
\email{srinath@wso2.com}
\orcid{0000-0002-4457-903X}

\renewcommand{\shortauthors}{Samaraweera et al.}

\begin{abstract}
Large Language Models achieve strong code generation for high resource languages like Python and Java but suffer sharp performance drops on Low-Resource Programming Languages~(LRPLs) such as Julia. Improving Small Language Models~(SLMs) for these languages faces a trilemma: Supervised Fine-Tuning~(SFT) is bottlenecked by data scarcity, inference-time scaling is too expensive for deployment, and Reinforcement Learning from scratch yields near zero advantages.
We propose a three-phase pipeline that resolves this trilemma by decoupling syntax acquisition from algorithmic reasoning. First, we \emph{left-shift} inference-time compute to an offline data synthesis engine that uses iterative compiler and test feedback to generate verified training examples. Second, we fine-tune an SLM on this synthetic, verified data to embed strong syntactic priors. Third, we apply Reinforcement Learning with Verifiable Reward~(RLVR) grounded by language-agnostic Input/Output tests, where the SFT prior constrains exploration away from syntax errors.
Applied to Qwen3-8B, our pipeline improves pass@1 by up to +7.6 points on MultiPL-E and +14.2 points on the Agnostics LiveCodeBench for Julia compared to SOTA results. Furthermore, the pipeline only used $\frac{1}{3}$ data and $\frac{1}{6}$ cost over the previous state-of-the-art. We further demonstrate that the pipeline generalizes to Ballerina achieving 49.7\% MultiPL-E Pass@1, a language with near-zero pretraining representation. Ablations confirm that both the SFT phase and execution-grounded rewards are necessary for stable training.
\end{abstract}




\maketitle


\section{Introduction}
\label{sec:intro}
Large Language Models (LLMs) have achieved remarkable proficiency in code generation for high-resource programming languages(HRPLs) like Python and Java~\cite{chen2021humaneval}. However, their performance sharply declines when applied to Low-Resource Programming Languages (LRPLs)~\cite{multipl-e, joel2024survey}. This disparity is not merely academic: languages such as Julia (scientific computing) and Ballerina (cloud-native microservices) serve critical roles across diverse software engineering domains. As the ecosystem of programming languages continues to grow, with new languages emerging to address specialized needs, enabling reliable code generation for LRPLs is an increasingly pressing challenge for the software engineering community, mirroring the historical trajectory of low-resource Natural Language Processing~\cite{ranathunga2023neural}.

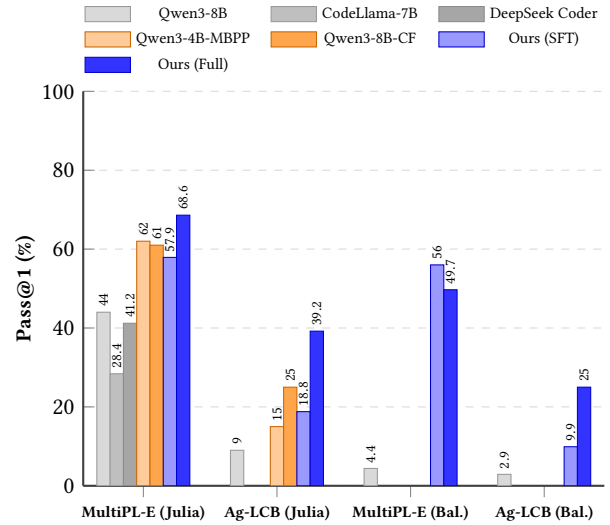
\begin{figure}[t]
    \centering
    \begin{tikzpicture}
        \begin{axis}[
            ybar=0pt, 
            area legend, 
            width=\columnwidth, 
            height=6.8cm,
            bar width=5pt, 
            enlarge x limits=0.15,
            ymin=0, ymax=100, 
            ylabel={\textbf{Pass@1 (\%)}},
            ylabel style={font=\small, yshift=-1ex},
            symbolic x coords={MultiPL-E (Julia), Ag-LCB (Julia), MultiPL-E (Bal.), Ag-LCB (Bal.)},
            xtick=data,
            xticklabel style={font=\scriptsize\bfseries},
            nodes near coords,
            every node near coord/.append style={rotate=90, anchor=west, font=\tiny, inner sep=2pt},
            axis line style={thick, darkgray},
            ymajorgrids=true,
            grid style={dashed, gray!30},
            axis x line*=bottom,
            axis y line*=left,
            legend style={
                at={(0.5, 1.02)},
                anchor=south,
                legend columns=3,
                font=\scriptsize,
                draw=none,
                fill=none,
                /tikz/every even column/.append style={column sep=0.2cm}
            }
        ]
        
        \addplot[fill=gray!30, draw=gray!80] coordinates {
            (MultiPL-E (Julia), 44.0) 
            (Ag-LCB (Julia), 9.0)
            (MultiPL-E (Bal.), 4.4) 
            (Ag-LCB (Bal.), 2.9)
        };
        \addlegendentry{Qwen3-8B}

        \addplot[fill=gray!50, draw=gray!80] coordinates {
            (MultiPL-E (Julia), 28.4) 
            (Ag-LCB (Julia), nan)
            (MultiPL-E (Bal.), nan) 
            (Ag-LCB (Bal.), nan)
        };
        \addlegendentry{CodeLlama-7B}

        \addplot[fill=gray!70, draw=gray!90] coordinates {
            (MultiPL-E (Julia), 41.2) 
            (Ag-LCB (Julia), nan)
            (MultiPL-E (Bal.), nan) 
            (Ag-LCB (Bal.), nan)
        };
        \addlegendentry{DeepSeek Coder}

        \addplot[fill=orange!40, draw=orange!80!black] coordinates {
            (MultiPL-E (Julia), 62.0) 
            (Ag-LCB (Julia), 15.0)
            (MultiPL-E (Bal.), nan) 
            (Ag-LCB (Bal.), nan)
        };
        \addlegendentry{Qwen3-4B-MBPP}

        \addplot[fill=orange!70, draw=orange!80!black] coordinates {
            (MultiPL-E (Julia), 61.0) 
            (Ag-LCB (Julia), 25.0)
            (MultiPL-E (Bal.), nan) 
            (Ag-LCB (Bal.), nan)
        };
        \addlegendentry{Qwen3-8B-CF}

        \addplot[fill=blue!40, draw=blue!80!black] coordinates {
            (MultiPL-E (Julia), 57.9) 
            (Ag-LCB (Julia), 18.8)
            (MultiPL-E (Bal.), 56.0) 
            (Ag-LCB (Bal.), 9.9)
        };
        \addlegendentry{Ours (SFT)}

        \addplot[fill=blue!80, draw=blue!80!black, nodes near coords style={font=\tiny\bfseries}] coordinates {
            (MultiPL-E (Julia), 68.6) 
            (Ag-LCB (Julia), 39.2)
            (MultiPL-E (Bal.), 49.7) 
            (Ag-LCB (Bal.), 25.0)
        };
        \addlegendentry{Ours (Full)}

        \end{axis}
    \end{tikzpicture}
    \caption{Comprehensive evaluation of code generation performance (Pass@1 \%). Our full pipeline improves over the Qwen3-8B base model across all splits, matching or substantially outperforming other baselines and prior state-of-the-art methodologies.}
    \Description{...}
    \label{fig:teaser_clustered_adjacent}
\end{figure}

Enhancing Small Language Models (SLMs) for LRPLs presents a trilemma of distinct challenges. \textbf{First}, Supervised Fine-Tuning (SFT) is highly effective for smaller parameter models~\cite{giagnorio2025enhancing}, but it is fundamentally bottlenecked by severe data scarcity.  For example, datasets like CodeNet~\cite{puri2021codenetlargescaleaicode} contain only 10.8k Julia samples compared to 3.2M for Python. \textbf{Second}, Test-time Scaling (\eg iterative self-correction, reflection-based refinement) effectively boosts performance~\cite{Wang2025CodeChemistFK, ren2024reflectioncoder}, but introduces unacceptable latency and computational overhead for real-world deployments. \textbf{Third}, applying Reinforcement Learning (RL) directly to a base model suffers from sparse rewards: e.g., the model generates so many syntax errors that it rarely receives \ref{fig:reward_std} positive feedback to learn from.

This paper uses on two key hypotheses to improve SLMs for LRPLs.

Our first hypothesis is that \emph{inference-time scaling and supervised fine-tuning are not competing strategies but can be complementary phases of a single pipeline}. Rather than deploying computationally expensive iterative refinement during live usage, we can ``left-shift'' this computation, repurposing it strictly as an offline data curation engine. By utilizing iterative compiler and test-case feedback, where compilation errors or a single failed test case showing the expected versus actual output are fed back to the model for self-correction, we synthesize missing ground-truth data to enable high-quality SFT. 

A central challenge in this pipeline is ensuring the quality of the synthetically generated data: an LLM can produce fluent but semantically incorrect code, and training on such examples would embed faulty priors. We address this through a flexible verification strategy adapted to the available resources. When a ground-truth solution exists in a high-resource language (\eg Python), we can employ \emph{differential testing}, executing both the candidate LRPL solution and the reference implementation on shared inputs and comparing their outputs. Differential testing is preferred when available because it catches subtle semantic divergences that fixed IO test suites may miss, such as floating-point precision differences, locale-dependent string formatting, or edge cases in boundary conditions that the original test suite does not explicitly cover. When no reference solution is available, we fall back to \emph{input/output test verification}, running the candidate against a suite of I/O test cases derived from the problem specification. In practice, the majority of our seed problems originate from competitive programming datasets that provide IO test suites but not reference solutions in the target LRPL, making IO verification the dominant verification path in our experiments. This two-tiered approach is deliberately pragmatic: the verification method is determined entirely by what artifacts are available for a given problem, requiring no additional human annotation.

Our second hypothesis is that the \textit{improvement achieved via Reinforcement Learning is dependent on the difficulty of the problems it gets feedback on}. With very easy problems, there is little to learn, and with very hard problems, the model's answers will all be wrong; thus, the model will not know how to make progress. Problems with the correct difficulty can speed up the learning. For example, Pikus et al. ~\cite{pikus2025hard} have also observed that hard samples help with GRPO learning. This idea can also be seen as a variation of active learning  (Ren et al.\cite{ren2021survey}), where we actively select the right data to be included in the training set to maximize the model outcome. 

Building on this idea, we create a dataset for Reinforcement Learning with Verifiable Rewards (RLVR) training by using difficulty scores (ELO rating for problems) to curate the data. Subsequently, we apply RLVR via Group Relative Policy Optimization (GRPO)~\cite{shao2024deepseekmath}, directly adopting the language-agnostic IO-based evaluation framework proposed by Boruch-Gruszecki et al.~\cite{boruch-gruszecki2026agnostics}. To maximize the learning signal, we employ partial rewards proportional to the fraction of test cases passed and successful compilation, allowing the advantage estimates to capture fine-grained differences between completions. 

As we will see in our results analysis, because the SFT prior constrains the action space away from trivial syntax errors, the RL phase can efficiently navigate the richer landscape of functional logic. Furthermore, to prevent the policy from converging to sub-optimal states, we use \emph{zero-advantage masking} \cite{Chen2025PanguEA}: if no completion within a group achieves the maximum reward, the entire group's advantage is set to zero, effectively discarding the update. This avoids reinforcing mediocre solutions and ensures that the policy only updates toward genuinely strong completions.

We evaluate our pipeline on Qwen3-8B \cite{qwen3} using two benchmarks: MultiPL-E~\cite{multipl-e} and the Agnostics LiveCodeBench (Ag-LCB) \cite{boruch-gruszecki2026agnostics}. For Julia, our full pipeline improves pass@1 by +24.6 points on MultiPL-E and +30.2 points on Ag-LCB. For Ballerina, where the base model has near-zero prior knowledge, we observe +45.3 points improvement on MultiPL-E and +22.1 points on Ag-LCB over the baseline. Notably, the SFT phase alone achieves on-par results with the current state-of-the-art, demonstrating that left-shifted inference-time compute can match expensive online scaling strategies through a one-time offline investment. Furthermore, validating our second hypothesis, we achieved the above results while using $\frac{1}{3}$ of the data and $\frac{1}{6}$ of the cost.

We further demonstrate that the pipeline generalizes to Ballerina, a cloud-native programming language for integration and microservices that has near-zero representation in LLM pretraining corpora. Unlike Julia, which benefits from some incidental coverage in training data, Ballerina presents a true cold-start scenario: the base model has effectively no prior exposure to its syntax, type system, or standard library. Despite this, our pipeline bootstraps functional Ballerina code generation by relying solely on compiler feedback and language-agnostic test suites, requiring no human-written Ballerina training examples. This result provides strong evidence that any LRPL can be bootstrapped making the pipeline immediately applicable as new programming languages emerge.

Our contributions are as follows:
\begin{itemize}
   \item We propose a three-phase pipeline, offline inference scaling via iterative compiler and test feedback, SFT, and GRPO with a dataset curated by difficulty, that achieves state-of-the-art results on MultiPL-E and Ag-LCB for Julia and Ballerina using an 8B-parameter model, with much fewer training rounds and lower cost.
  \item We demonstrate that left-shifting inference-time computation to offline data synthesis resolves the data scarcity bottleneck without human annotation, turning expensive iterative refinement into a one-time investment.
  \item We show that the pipeline generalizes to an extreme LRPL (Ballerina) with minimal adaptation, establishing a practical recipe for bootstrapping code generation for any LRPL.
  \item We extend the MultiPL-E benchmark for Ballerina, adding a new language to the evaluation suite and providing a standardized testbed for future research on LRPL code generation.

\end{itemize}


\section{Related Work}
\label{sec:related}

Benchmarking LLMs on LRPLs relies on frameworks adapted from high-resource environments, such as MultiPL-E~\cite{multipl-e} and Agnostic-LiveCodeBench \cite{jain2024livecodebench}. While these benchmarks effectively measure performance, improving that performance requires navigating severe data limitations and choosing optimal learning strategies. We organize prior work along three axes that correspond to the phases of our pipeline.

\subsection{SFT and the Challenge of Data Scarcity}

Supervised Fine-Tuning remains the most prevalent strategy for adapting models to LRPLs. Cassano et al.~\cite{cassano:multipl-t} introduced MultiPL-T, a pipeline for translating training data from high-resource to low-resource languages, demonstrating significant gains for models up to 8B parameters. Giagnorio et al.~\cite{giagnorio2025enhancing} further showed that smaller models ($\sim$1B parameters) trained on specialized, domain-specific data often outperform larger, general-purpose models, as SFT can inversely degrade larger models better suited for Retrieval-Augmented Generation. WizardCoder~\cite{luo2024wizardcoder} and OpenCoder~\cite{Huang2024OpenCoderTO} advanced SFT through instruction-tuning and open-source training recipes, though their focus remains on high-resource languages. Joel et al.~\cite{joel2024survey} provide a comprehensive survey of LLM-based code generation for low-resource and domain-specific languages.

The fundamental bottleneck for SFT is data scarcity. Datasets like CodeNet~\cite{puri2021codenetlargescaleaicode} provide a foundation, but the available data for LRPLs pales in comparison to high-resource languages. Utilizing organic data requires rigorous curation, including deduplication and strict decontamination to prevent benchmark leakage. To bypass organic data limitations, recent research emphasizes synthetic data pipelines: Magicoder~\cite{wei2024magicoder} leverages OSS-Instruct to generate training data from open-source snippets, while KODCODE~\cite{xu2025kodcode} constructs a diverse, verifiable synthetic dataset. Our work extends this line by introducing \emph{compiler/test feedback driven} data synthesis, where iterative execution and error correction provides verification that purely generative pipelines lack.

\subsection{Reinforcement Learning and Grounding}

Beyond SFT, Reinforcement Learning is establishing new state-of-the-art results for code generation. Early approaches used Proximal Policy Optimization (PPO): CodeRL~\cite{le2022coderl} combined pretrained models with execution-based RL rewards, while Shojaee et al.~\cite{shojaee2023executionbased} demonstrated that execution-based feedback outperforms static heuristics for policy optimization. More recently, Group Relative Policy Optimization (GRPO)~\cite{shao2024deepseekmath} has emerged as a simpler alternative that eliminates the need for a separate value network. CodeReasoner~\cite{tang2025codereasonerenhancingcodereasoning} applies GRPO to enhance code reasoning in general-purpose models.

For LRPLs specifically, recent techniques rely on language agnostic datasets and test-feedback loops. Concurrently, self-play frameworks enable iterative refinement without human annotation \cite{lin2025learning}. However, ungrounded self-play often yields sub-optimal results. To address this problem, Boruch-Gruszecki et al. \cite{boruch-gruszecki2026agnostics} introduced Agnostics, a universal RL environment capable of learning to synthesize code in any programming language via execution-based rewards. 

Ma et al. \cite{ma2026learning} demonstrate that interleaving online fine-tuning with RL addresses the hardest questions that reinforcement learning alone cannot solve, highlighting the complementary nature of supervised and RL-based training.

Our work grounds RL with IO tests similar to Boruch-Gruszecki et al.\ and, crucially, initializes the RL policy from a \emph{syntax-primed} SFT model rather than the base model, which dramatically reduces the sparse-reward problem.

\subsection{Test-time Scaling and Its Limitations}

To compensate for limited training data, the current state-of-the-art often allocates more computational resources during execution, a paradigm known as Test-time Scaling. A prominent family of techniques uses \emph{iterative refinement}: the model generates a candidate solution, receives feedback (compiler errors, test failures, or self-reflection), and refines its output across multiple rounds. ReflectionCoder~\cite{ren2024reflectioncoder} leverages reflection sequences to iteratively correct one-off generations, while Origen~\cite{cui2024origen} uses code-to-code augmentation with self-reflection for RTL code generation. AgentCoder~\cite{huang2023agentcoder} employs a multi-agent framework where a programmer agent, test designer agent, and test executor agent collaborate through iterative testing and optimization. CodeSim~\cite{codesim2025} extends this paradigm with simulation-driven planning and debugging.

Another family uses \emph{selection-based scaling}: CodeT~\cite{Chen2022CodeTCG} generates multiple candidate solutions along with tests and selects the best via dual execution agreement. CodeChemist~\cite{Wang2025CodeChemistFK} applies functional knowledge transfer with test-time scaling specifically for low-resource code generation.

While these methods form a critical pillar of modern code generation, their high computational cost at inference time makes them impractical for efficient SLM deployment. Our work reframes iterative refinement, the most effective form of inference-time scaling, not as a deployment strategy but as an \emph{offline data curation} mechanism, amortizing the compute cost across all future queries.

\section{Methodology}
\label{sec:method}

\subsection{Pipeline Overview}
\label{sec:method:overview}

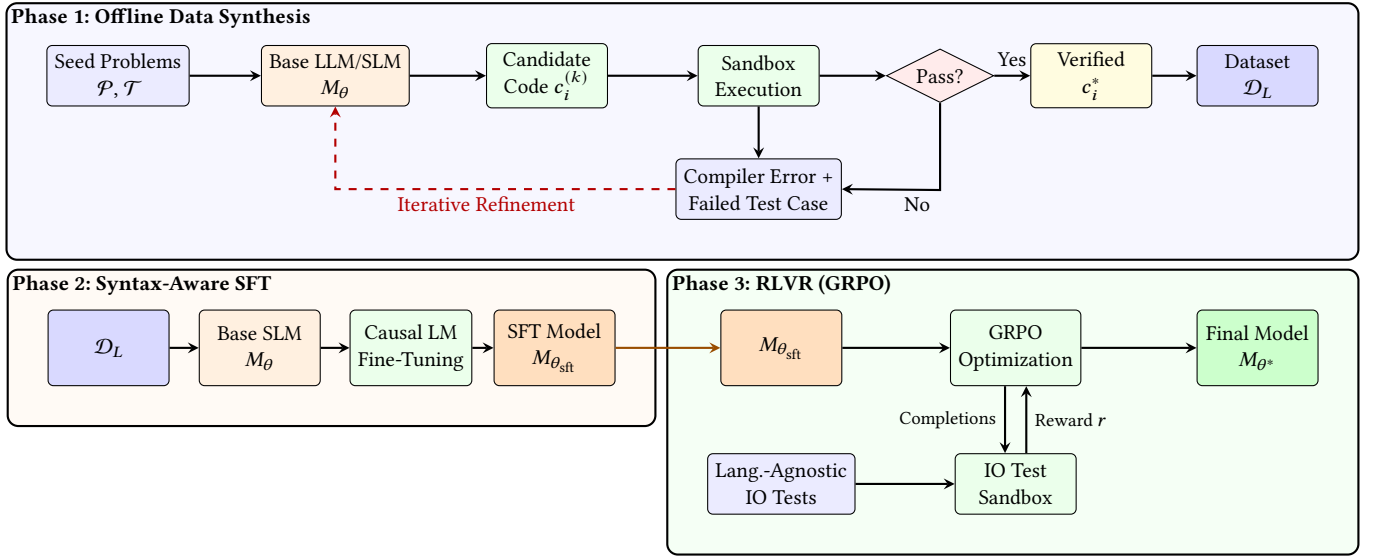
\begin{figure*}[t]
  \centering
  \begin{tikzpicture}[
    node distance=0.5cm and 0.7cm,
    box/.style={draw, rounded corners=2pt, minimum height=0.8cm, minimum width=1.6cm,
                align=center, font=\small, inner sep=3pt},
    phasebox/.style={draw, rounded corners=4pt, inner sep=18pt, thick},
    phaselabel/.style={font=\small\bfseries, fill=none, inner sep=2pt},
    arrow/.style={->, >=stealth, thick},
    feedbackarrow/.style={->, >=stealth, thick, dashed, red!70!black},
    data/.style={box, fill=blue!8},
    model/.style={box, fill=orange!12},
    process/.style={box, fill=green!8},
    output/.style={box, fill=yellow!15},
    decision/.style={draw, diamond, aspect=2, fill=red!8, minimum width=0.8cm,
                     minimum height=0.6cm, align=center, font=\small, inner sep=2pt},
  ]


    \node[data] (seeds) at (0.125, 0) {Seed Problems\\$\mathcal{P}, \mathcal{T}$};
    \node[model] (baseslm1) at (3.0, 0) {Base LLM/SLM\\$M_\theta$};
    \node[process] (candidate) at (5.8, 0) {Candidate\\Code $c_i^{(k)}$};
    \node[process] (sandbox1) at (8.6, 0) {Sandbox\\Execution};
    \node[decision] (pass) at (11.0, 0) {\small Pass?};
    \node[output] (verified) at (13.0, 0) {Verified\\$c_i^*$};
    \node[output, fill=blue!15] (dataset) at (15.2, 0) {Dataset\\$\mathcal{D}_L$};

    \node[data] (feedback) at (8.6, -1.5) {Compiler Error +\\Failed Test Case};

    \draw[arrow] (seeds) -- (baseslm1);
    \draw[arrow] (baseslm1) -- (candidate);
    \draw[arrow] (candidate) -- (sandbox1);
    \draw[arrow] (sandbox1) -- (pass);
    \draw[arrow] (pass) -- node[above, font=\small] {Yes} (verified);
    \draw[arrow] (verified) -- (dataset);
    \draw[arrow] (sandbox1) -- (feedback);
    \draw[feedbackarrow] (feedback.west) -| (baseslm1.south);
    \node[font=\small, red!70!black] at (5.0, -1.7) {Iterative Refinement};
    \draw[arrow] (pass.south) -- ++(0, -1.15cm) -- (feedback.east);
    \node[font=\small] at (10.7, -1.7) {No};

    \begin{scope}[on background layer]
      \node[phasebox, fill=blue!3, fit=(seeds)(baseslm1)(candidate)(sandbox1)(pass)(verified)(dataset)(feedback),
            inner sep=15pt] (phase1box) {};
    \end{scope}
    \node[phaselabel, anchor=north west] at (phase1box.north west) {Phase 1: Offline Data Synthesis};

    \node[output, fill=blue!15, minimum height=1.0cm] (dataset2) at (0, -3.6) {$\mathcal{D}_L$};
    \node[model, minimum height=1.0cm] (baseslm2) at (2.0, -3.6) {Base SLM\\$M_\theta$};
    \node[process, minimum height=1.0cm] (sftproc) at (4.0, -3.6) {Causal LM\\Fine-Tuning};
    \node[model, fill=orange!25, minimum height=1.0cm] (sftmodel) at (5.9, -3.6) {SFT Model\\$M_{\theta_{\text{sft}}}$};

    \draw[arrow] (dataset2) -- (baseslm2);
    \draw[arrow] (baseslm2) -- (sftproc);
    \draw[arrow] (sftproc) -- (sftmodel);

    \coordinate (p2bottom) at (-0.5, -4.0);
    \begin{scope}[on background layer]
      \node[phasebox, fill=orange!4, fit=(dataset2)(baseslm2)(sftproc)(sftmodel)(p2bottom),
            inner sep=15pt] (phase2box) {};
    \end{scope}
    \node[phaselabel, anchor=north west] at (phase2box.north west) {Phase 2: Syntax-Aware SFT};

    \node[model, fill=orange!25, minimum height=1.0cm] (sftinit) at (8.9, -3.6) {$M_{\theta_{\text{sft}}}$};
    \node[process, minimum height=1.0cm] (grpo) at (12.0, -3.6) {GRPO\\Optimization};
    \node[model, fill=green!20, minimum height=1.0cm] (finalmodel) at (15.2, -3.6) {Final Model\\$M_{\theta^*}$};
    \node[process, minimum height=0.8cm] (sandbox2) at (12.0, -5.4) {IO Test\\Sandbox};
    \node[data, minimum height=0.8cm] (iotests) at (8.9, -5.4) {Lang.-Agnostic\\IO Tests};

    \draw[arrow] (sftinit) -- (grpo);
    \draw[arrow] (grpo) -- (finalmodel);
    \draw[arrow] (iotests) -- (sandbox2);
    \draw[arrow] ([xshift=-4pt]grpo.south) -- ([xshift=-4pt]sandbox2.north)
      node[midway, left, font=\footnotesize] {Completions};
    \draw[arrow] ([xshift=4pt]sandbox2.north) -- ([xshift=4pt]grpo.south)
      node[midway, right, font=\footnotesize] {Reward $r$};

    \draw[arrow, thick, orange!60!black] (sftmodel) -- (sftinit);

    \begin{scope}[on background layer]
      \node[phasebox, fill=green!4, fit=(sftinit)(grpo)(sandbox2)(iotests)(finalmodel),
            inner sep=15pt] (phase3box) {};
    \end{scope}
    \node[phaselabel, anchor=north west] at (phase3box.north west) {Phase 3: RLVR (GRPO)};

  \end{tikzpicture}
  \caption{Overview of the three-phase pipeline. Phase~1 repurposes iterative refinement as an offline data curation engine: compiler errors and failed test cases (with expected vs.\ actual output) are fed back to the model for self-correction until all tests pass. Phase~2 fine-tunes the base model on synthesized data to embed syntax priors. Phase~3 applies GRPO with language-agnostic IO tests, exploiting the constrained action space from SFT for efficient RL exploration. \textbf{Takeaway:} Expensive inference compute is invested once offline, yielding a deployment-efficient SLM.}
  \Description{...}
  \label{fig:pipeline}
\end{figure*}

Our pipeline consists of three sequential phases, each addressing a distinct aspect of the LRPL code generation challenge. Figure~\ref{fig:pipeline} provides an architectural overview.

Given a target low-resource language $L$, a set of language independent seed problems $\mathcal{P} = \{p_1, \ldots, p_N\}$ with corresponding test suites $\mathcal{T} = \{t_1, \ldots, t_N\}$, and a pretrained base model $M_\theta$, our goal is to produce a fine-tuned model $M_{\theta^*}$ that maximizes pass@1 on held-out problems in $L$. The pipeline proceeds as follows: (1)~\textbf{Data Synthesis} uses $M_\theta$ with iterative compiler feedback to construct a verified dataset $\mathcal{D}_L$; (2)~\textbf{Syntax-Aware SFT} fine-tunes $M_\theta$ on $\mathcal{D}_L$ to produce $M_{\theta_{\text{sft}}}$ with strong syntactic priors; (3)~\textbf{RLVR} applies GRPO starting from $M_{\theta_{\text{sft}}}$ to optimize functional correctness, producing the final model $M_{\theta^*}$.

\subsection{Phase 1: Data Synthesis via Offline Inference Scaling}
\label{sec:method:synthesis}

To overcome the LRPL data bottleneck, we repurpose expensive inference-time scaling, specifically iterative self-correction with compiler/test-case feedback, strictly as an automated, offline data curation engine.

\subsubsection{Problem Formulation.}
Let $\mathcal{P} = \{p_1, \ldots, p_N\}$ be a set of language independent algorithmic problems, each paired with a test suite $t_i \in \mathcal{T}$ consisting of IO pairs. For a target language $L$, we seek to construct a supervised dataset:
\begin{equation}
  \mathcal{D}_L = \{(p_i, c_i^*) \mid c_i^* \text{ passes all tests in } t_i \text{ when executed in } L\}
  \label{eq:dataset}
\end{equation}

\subsubsection{Iterative Compiler/Test Feedback Refinement.}
For each problem $p_i$, the base model $M_\theta$ generates an initial candidate solution $c_i^{(0)}$. This candidate is executed in an isolated sandbox against the test suite $t_i$, yielding either success or a feedback signal $e_i^{(0)}$. The feedback comprises either one from two components: (1)~any compilation or runtime error message, or in case of compilation success and test cases are failing (2)~a single failed test case showing the expected output alongside the actual output produced by the candidate. If a candidate fails, we inject this structured feedback back into the model's context, prompting iterative refinement:
\begin{equation}
  c_i^{(k+1)} = M_\theta\!\left(p_i,\; c_i^{(k)},\; e_i^{(k)}\right), \quad k = 0, 1, \ldots, K_{\max}
  \label{eq:refinement}
\end{equation}
The loop terminates when $\mathrm{pass}(c_i^{(k)}, t_i) = 1$ (all tests pass) or $k > K_{\max}$. Successful trajectories contribute the final solution $c_i^* = c_i^{(k)}$ to $\mathcal{D}_L$; failed trajectories are discarded. This process is identical to iterative refinement used at inference time by approaches like ReflectionCoder~\cite{ren2024reflectioncoder} and AgentCoder~\cite{huang2023agentcoder}, but we perform it \emph{once, offline}, and retain only the final verified solutions as training data.

\textbf{Algorithm~\ref{alg:synthesis}} summarizes the complete data synthesis procedure.

\begin{algorithm}[t]
\caption{Offline Data Synthesis via Compiler/Test-Case Feedback}
\label{alg:synthesis}
\begin{algorithmic}[1]
\REQUIRE Seed problems $\mathcal{P}$, test suites $\mathcal{T}$, base model $M_\theta$, max iterations $K_{\max}$, target language $L$
\ENSURE Verified dataset $\mathcal{D}_L$
\STATE $\mathcal{D}_L \leftarrow \emptyset$
\FOR{each $(p_i, t_i) \in (\mathcal{P}, \mathcal{T})$}
    \STATE $c_i^{(0)} \leftarrow M_\theta(p_i, L)$ \COMMENT{Initial generation}
    \FOR{$k = 0$ \TO $K_{\max}$}
        \STATE $(s, e_i^{(k)}) \leftarrow \textsc{SandboxExec}(c_i^{(k)}, t_i, L)$
        \IF{$s = \texttt{PASS}$}
            \STATE $\mathcal{D}_L \leftarrow \mathcal{D}_L \cup \{(p_i, c_i^{(k)})\}$
            \STATE \textbf{break}
        \ENDIF
        \STATE $c_i^{(k+1)} \leftarrow M_\theta(p_i, c_i^{(k)}, e_i^{(k)})$ \COMMENT{Feedback-guided refinement}
    \ENDFOR
\ENDFOR
\RETURN $\mathcal{D}_L$
\end{algorithmic}
\end{algorithm}

\subsubsection{Seed Selection.}
We initialize the pipeline with language independent algorithmic problems sourced from [dataset]. These problems are specified via natural-language descriptions and IO test pairs, ensuring no dependence on any specific programming language. 

\subsubsection{Generation Prompt.}
Each problem is presented to the base model $M_\theta$ using a structured system prompt that constrains the output format and encourages reasoning before coding. Key design decisions include: (1)~requiring step-by-step reasoning in comments before the solution, mirroring chain-of-thought prompting; (2)~enforcing a strict output structure with separate \texttt{<CODE>} and \texttt{<TESTS>} blocks to facilitate automated extraction and validation; (3)~mandating a \texttt{main()} function that reads from standard input, enabling execution against IO test suites; and (4)~requesting at least four unit test assertions per problem to provide additional verification signal.





\subsubsection{Worked Example.}
Figure~\ref{fig:worked_example} illustrates the iterative refinement process on a concrete Julia problem: writing a function to sum the digits of a number. The base model's initial attempt produces syntactically valid Julia but fails all four test cases, a common pattern where models transfer Python semantics (e.g., integer division behavior or string-conversion idioms) that do not generalize to Julia. The sandbox feeds back the test failure summary (\texttt{0 passed, 4 failed}), which is appended to the prompt. In the second iteration, the model self-corrects by using Julia's idiomatic \texttt{string}/\texttt{abs}/character-arithmetic approach, producing a solution that passes all tests. This single-retry success demonstrates that even one round of compiler-and-test feedback is often sufficient to bridge the gap between high-resource habits and LRPL-correct code. The final, verified solution is added to $\mathcal{D}_L$.

\begin{figure}[t]
  \centering
  \small
  \setlength{\fboxsep}{6pt}
  \begin{tikzpicture}[
    iterbox/.style={draw, rounded corners=3pt, fill=#1, text width=0.88\columnwidth, inner sep=6pt, font=\small},
    iterbox/.default={white},
    arrowlabel/.style={font=\small\itshape, text=red!70!black, align=center},
  ]

    \node[iterbox=blue!5, draw=blue!40] (problem) at (0, 0) {
      \textbf{Problem:} \textit{``Write a function to calculate the sum of the digits in a given number.''}\\[2pt]
    };

    \node[iterbox=red!4, draw=red!30, below=0.35cm of problem] (iter0) {
      \textbf{Iteration 1} \hfill \colorbox{red!12}{\scriptsize\strut\textcolor{red!70!black}{~Test Failure~}}\\[3pt]
      \texttt{function sum\_digits(n::Integer)}\\[-1pt]
      \texttt{~~~~total = 0}\\[-1pt]
      \texttt{~~~~while n > 0~~\textcolor{red!70!black}{\# fails for negative input}}\\[-1pt]
      \texttt{~~~~~~~~total += n \% 10}\\[-1pt]
      \texttt{~~~~~~~~n = div(n, 10)}\\[-1pt]
      \texttt{~~~~end}\\[-1pt]
      \texttt{~~~~return total}\\[-1pt]
      \texttt{end}\\[2pt]
      {\small$\boldsymbol{\times}$}~\textit{\textcolor{red!70!black}{0 passed, 4 failed}}
    };

    \draw[-stealth, thick, red!60!black] (iter0.south) -- ++(0, -0.5cm);

    \node[iterbox=green!5, draw=green!50!black, below=0.6cm of iter0] (iter1) {
      \textbf{Iteration 2} \hfill \colorbox{green!15}{\scriptsize\strut\textcolor{green!50!black}{~Correct~$\to$~added to $\mathcal{D}_L$~}}\\[3pt]
      \texttt{function sum\_of\_digits(n::Integer)}\\[-1pt]
      \texttt{~~~~s = string(abs(n))}\\[-1pt]
      \texttt{~~~~return sum(c - '0' for c in s)}\\[-1pt]
      \texttt{end}\\[3pt]
      \texttt{\textcolor{green!50!black}{using Test}}\\[-1pt]
      \texttt{\textcolor{green!50!black}{@testset "Sum of digits" begin}}\\[-1pt]
      \texttt{\textcolor{green!50!black}{~~~~@test sum\_of\_digits(123) == 6}}\\[-1pt]
      \texttt{\textcolor{green!50!black}{~~~~@test sum\_of\_digits(-456) == 15}}\\[-1pt]
      \texttt{\textcolor{green!50!black}{~~~~@test sum\_of\_digits(0) == 0}}\\[-1pt]
      \texttt{\textcolor{green!50!black}{~~~~@test sum\_of\_digits(1000) == 1}}\\[-1pt]
      \texttt{\textcolor{green!50!black}{end}}\\[2pt]
      {\normalsize$\checkmark$}~\textit{\textcolor{green!50!black}{All 4 tests pass}}
    };

  \end{tikzpicture}
  \caption{Worked example of iterative compiler-feedback refinement for Julia. The base model's first attempt compiles but fails all tests due to incorrect handling of negative numbers and modular arithmetic. After one round of test feedback, the model self-corrects to an idiomatic Julia solution using \texttt{abs} and character arithmetic. Only the verified solution enters $\mathcal{D}_L$.}
  \Description{...}
  \label{fig:worked_example}
\end{figure}


\subsection{Phase 2: Syntax-Aware Supervised Fine-Tuning}
\label{sec:method:sft}

Using the synthesized dataset $\mathcal{D}_L$, we fine-tune the pretrained base model $M_\theta$ using standard causal language modeling. The SFT objective minimizes the cross-entropy loss over verified prompt-completion pairs:
\begin{equation}
  \mathcal{L}_{\text{SFT}} = -\sum_{(p, c^*) \in \mathcal{D}_L} \sum_{t=1}^{|c^*|} \log \pi_\theta(c^*_t \mid p, c^*_{<t})
  \label{eq:sft}
\end{equation}
where $\pi_\theta$ denotes the model's output distribution.

This phase serves a specific purpose: because LRPLs possess idiosyncratic syntaxes and niche standard libraries that differ substantially from high-resource languages, SFT embeds strong structural and grammatical priors into the model. Consider the differences between Julia and Python that a model must internalize: Julia uses 1-based indexing, \texttt{true}/\texttt{false} instead of \texttt{True}/\texttt{False}, \texttt{function}/\texttt{end} blocks instead of indentation-delimited scopes, multiple dispatch instead of class-based OOP, and a distinct standard library (e.g., \texttt{sort(arr, rev=true)} vs.\ \texttt{sorted(arr, reverse=True)}). After this phase, the model can reliably produce syntactically valid Julia code structures before being required to optimize for complex algorithmic logic in the subsequent RL phase.

For Ballerina, the syntactic gap is even larger: the language uses a unique type system with union types (\texttt{int|string}), built-in concurrency via workers, explicit error handling with \texttt{check} expressions, and a service-oriented programming model unlike any high-resource language. The SFT phase is therefore \emph{essential} for Ballerina, as the base model has effectively zero prior exposure to these constructs.

In practice, each training example in $\mathcal{D}_L$ is formatted as a prompt-completion pair: the prompt contains the natural-language problem description along with a language-specific instruction header (\eg ``Write a Julia function that...''), and the completion is the verified solution from Phase~1. During fine-tuning, we apply \emph{completion-only loss masking}, computing the cross-entropy loss exclusively over the solution tokens rather than the prompt tokens, which focuses the learning signal on code generation rather than problem comprehension. 


\subsection{Phase 3: RLVR with Verifiable Rewards with Difficulty-curated Data}
\label{sec:method:rl}

\subsubsection{Data Curation by Difficulty}

With a robust syntax prior established, we transition to exploration via Reinforcement Learning. Crucially, we initialize the GRPO policy with the syntax-aware SFT model $M_{\theta_{\text{sft}}}$ from Phase~2, rather than the base model $M_\theta$.

\subsubsection{GRPO Objective.}
For a given problem $p$, we sample a group of $G$ candidate completions $\{c_1, \ldots, c_G\}$ from the current policy $\pi_\theta$. Each completion receives a reward $r_i$ from the execution-based oracle. We compute group-relative advantages:
\begin{equation}
  \hat{A}_i = \frac{r_i - \mu_G}{\sigma_G + \epsilon}
  \label{eq:advantage}
\end{equation}
where $\mu_G$ and $\sigma_G$ are the mean and standard deviation of rewards within the group, and $\epsilon$ is a small constant for numerical stability. The policy is updated to maximize:
\begin{equation}
  \mathcal{L}_{\text{GRPO}} = \mathbb{E}_{p \sim \mathcal{P}} \left[ \frac{1}{G} \sum_{i=1}^{G} \hat{A}_i \cdot \log \pi_\theta(c_i \mid p) \right] - \beta \cdot D_{\text{KL}}\!\left(\pi_\theta \| \pi_{\text{ref}}\right)
  \label{eq:grpo}
\end{equation}
where $\pi_{\text{ref}} = \pi_{\theta_{\text{sft}}}$ is the SFT model serving as the reference policy, and $\beta$ controls the strength of the KL divergence penalty that prevents the policy from deviating too far from the syntactic prior.

\subsubsection{Reward Function.}
\label{reward-function}
Each completion $c$ receives a composite reward that decomposes into a \emph{test score} $r_{\text{test}}$ and a \emph{build-quality score} $r_{\text{build}}$, reflecting the two distinct failure modes in LRPL code generation: syntactic/compilation failures and logical/algorithmic failures.

The test score is proportional to the number of passed test cases $p$:
\begin{equation}
  r_{\text{test}}(c, p) = \min\!\left(\frac{p}{|\mathcal{T}_p|},\; 1.0\right)
  \label{eq:reward_test}
\end{equation}
This partial reward provides a gradient signal even for completions that solve only a subset of test cases, avoiding the sparsity of binary pass/fail rewards.

The build-quality score $r_{\text{build}}$ captures whether the completion compiles and produces well-structured code. If the code fails to compile, $r_{\text{build}}$ receives a small positive value inversely proportional to the number of compiler errors, rewarding completions that are ``closer'' to compiling. If compilation succeeds, $r_{\text{build}}$ receives a fixed credit regardless of test outcomes. Additionally, a \emph{structure penalty} is applied based on the presence of expected code constructs (function definitions, test blocks, assertions), acting as a negative offset that is reduced for well-formed outputs. This ensures that structural compliance alone cannot produce a high reward, only test-passing code is strongly reinforced. The build-quality score is down-weighted relative to the test score, ensuring that functional correctness dominates the learning signal:
\begin{equation}
  r(c, p) = r_{\text{test}}(c, p) + \alpha \cdot r_{\text{build}}(c, p)
  \label{eq:reward}
\end{equation}
where $\alpha < 1$ controls the relative weight of the build-quality shaping signal.

\subsubsection{Zero-Advantage Masking.}
A key challenge in GRPO for LRPLs is that groups may contain no meaningfully correct completion, all candidates fail most tests, producing near-uniform low rewards. Naively computing advantages in such groups reinforces the ``least bad'' completion, causing the policy to converge on sub-optimal heuristics (\eg hardcoding common outputs). We address this via \emph{zero-advantage masking}: if no completion within a group achieves the maximum reward, we set $\hat{A}_i = 0$ for all $i$ in that group, effectively discarding the update. This ensures the policy only learns from groups containing at least one genuinely strong completion, preventing noisy gradients from degrading the SFT prior.

Formally, let $r_{\max} = \max_{i \in G} (r_i)$ denote the highest reward in a group. We define the masked advantage as:
\begin{equation}
  \tilde{A}_i =
  \begin{cases}
    \hat{A}_i & \text{if } r_{\max} = 1.0 \;\text{(at least one fully correct completion)}, \\
    0         & \text{otherwise}.
  \end{cases}
  \label{eq:masked_advantage}
\end{equation}
Substituting $\tilde{A}_i$ into Eq.~\ref{eq:grpo} yields the masked GRPO objective:
\begin{equation}
  \mathcal{L}_{\text{GRPO}}^{\text{masked}} = \mathbb{E}_{p \sim \mathcal{P}} \left[ \frac{1}{G} \sum_{i=1}^{G} \tilde{A}_i \cdot \log \pi_\theta(c_i \mid p) \right] - \beta \cdot D_{\text{KL}}\!\left(\pi_\theta \| \pi_{\text{ref}}\right)
  \label{eq:grpo_masked}
\end{equation}
When $r_{\max} < 1.0$, the policy gradient term vanishes entirely for that group, since we set $\beta = 0$ similar to DAPO\cite{Yu2025DAPOAO} the entire sample is ignored.

\subsubsection{Data Curation by Difficulty}
\label{data-curation}

As discussed in the introduction, we aim to create a curated dataset for GRPO that can maximize learning. 

We use the ELO rating used by the Codeforces platform (see Quan et al.~\cite{quan2025codeelo} to measure problem difficulty, where both humans, models, and problems are given ELO ratings, which get adjusted using outcomes from humans and models attempting problems. Using the problem and ELO ratings data available from the Codeforces platform, we have built a model to predict the ELO rating given a problem. 

For GRPO, we use data that has IO-based ground truth as discussed by Agnostics paper, where we convert ground truth examples to test cases and then give feedback to GRPO using those tests as described in the subsection~\ref{reward-function}, 

Given reward function $r_{test}$, model $\ phi _ {ref} $, and a dataset $D$, that comprises a collection of problems, the ELO rating for the problem, and groundtruth data, we build a GRPO dataset as follows. 

\begin{equation}
  ELO_m = \min \left\{ elo \;\middle|\; \frac{dr_{test}}{delo} < -\theta \right\} \text{for a threshold theta}
  \label{eq:model-elo}
\end{equation}

This defines $ELO_m$ as the supremum (the highest point) of problem difficulty where the model maintains a stable reward:

When $d.elo$ is the elo rating of problem $d$, curated dataset is defined as follows. 

\begin{equation}
  D_{curated} = \{ d \in D s.t. ELO_m \leq d.elo \leq ELO_{m} + 400D.elo \} 
  \label{eq:d-curated}
\end{equation}

\subsubsection{Why SFT Initialization Matters.}
The interaction between Phase~2 and Phase~3 is the central design choice of our pipeline. GRPO works by generating candidate solutions, computing relative advantages from execution feedback, and adjusting the output distribution to favor better solutions. However, if all solutions within a group receive the same reward (\eg all compile but no test passes), the advantages collapse to zero and GRPO cannot learn. When GRPO is initialized from the base model without an SFT prior, the vast majority of sampled completions contain syntax errors and receive near-zero rewards dominated by the build-quality component. With no test-passing signal to differentiate completions, the advantages within each group lack meaningful variance and the policy stagnates. In contrast, the SFT-initialized policy produces syntactically valid completions that compile and execute, even if they produce wrong outputs. This means the reward distribution within each group has meaningful variance, some completions pass more tests than others, providing a rich gradient signal. This enables GRPO to learn effectively, and the compute budget is thus spent exploring functional logic rather than rediscovering basic language syntax.



\section{Experimental Setup}
\label{sec:experiments}

\subsection{Base Model}

We use Qwen3-8B~\cite{qwen3} as our base SLM. We select this model for four reasons: (1)~it demonstrates strong baseline performance on high-resource code generation, providing a capable starting point; (2)~at 8B parameters, it represents a realistic model size for on-premise deployment on a single GPU; (3)~its training data cutoff aligns with the LiveCodeBench v5 temporal split, enabling contamination-free evaluation on Ag-LCB; and (4) supports a fair comparison with the previous work, which reported against the same model. While more widely adopted recent models such as the Qwen3.5 family and DeepSeek-V3.2 series demonstrate stronger performance on high-resource languages, their later training cutoffs risk data contamination against our evaluation benchmarks.



\subsection{Target Languages}

We evaluate on two languages that represent different points on the resource spectrum:

\textbf{Julia} is designed for high-performance scientific computing. Despite a growing community, Julia remains underrepresented in LLM training corpora, CodeNet contains only 10.8k Julia samples versus 3.2M for Python. Julia's syntax combines familiar imperative structures with distinctive features: 1-based array indexing, multiple dispatch as the core paradigm, \texttt{function}/\texttt{end} block delimiters, and a rich type system with parametric types. These features make Julia a representative ``moderate LRPL'', present in pretraining data but insufficiently represented for reliable code generation.

\textbf{Ballerina} is a cloud-native programming language for building distributed applications and microservices. It features a unique type system with structural typing, union types (\texttt{int|string|error}), built-in concurrency via workers, first-class support for network services, and explicit error handling with \texttt{check} expressions. Ballerina has near-zero representation in LLM pretraining corpora, making it our ``extreme LRPL'' test case, a language where the base model has effectively no prior knowledge of the syntax or standard library.

\subsection{Datasets}
We source seed problems from the Stanford Alpaca dataset~\cite{alpaca}, selecting language-independent algorithmic tasks. For each problem and target language, we generate candidate solutions using Claude Code or OpenAI OSS-120B based on it's difficulty, then verify each candidate by executing its accompanying test suite. Only problems whose generated solutions pass all test cases are retained in the SFT dataset $\mathcal{D}_L$. This process is applied identically for both Julia and Ballerina, yielding fewer than 1{,}000 verified examples per language.

For the RL phase, we curate approximately 500 training problems from Codeforces (filtered to $\leq$1{,}300 ELO) and CodeNet~\cite{puri2021codenetlargescaleaicode}. Each problem is paired with language-agnostic IO test suites that serve as the verifiable reward signal during GRPO training.

\subsection{Benchmarks}

\subsubsection{MultiPL-E}~\cite{multipl-e} has translated HumanEval~\cite{chen2021humaneval} problems to 18+ programming languages, enabling standardized cross-lingual evaluation. We report pass@1 on the Julia split. As part of this work, we extend MultiPL-E to include Ballerina by manually translating 159 HumanEval problems into idiomatic Ballerina with corresponding test harnesses. Unlike automated translation approaches, our extension is entirely human-annotated to ensure correctness and idiomatic usage of Ballerina's distinctive features (\eg structural typing, \texttt{check} expressions, union types). This constitutes the first standardized code generation benchmark for Ballerina, which we release to support future research on emerging LRPLs. MultiPL-E problems are relatively straightforward, testing basic language competence and standard library knowledge.

\subsubsection{Ag-LCB}~\cite{boruch-gruszecki2026agnostics} is the LiveCodeBench~\cite{jain2024livecodebench} evaluation adapted through the Agnostics framework, using problems published after the base model's training cutoff to ensure contamination-free assessment. These problems are substantially harder than MultiPL-E, reflecting competitive programming difficulty with complex algorithmic reasoning requirements. We report pass@1 for both Julia and Ballerina.


\subsection{Baselines}

We compare the following configurations to isolate the contribution of each pipeline component:

\begin{itemize}
  \item \textbf{Qwen3-8B (Base)}: The unmodified pretrained model, establishing the baseline performance.
  \item \textbf{+ SFT Only}: Phase~2 only, the base model fine-tuned on our compiler-feedback synthetic data, without the subsequent RL phase. This isolates the contribution of syntax-aware SFT.
  \item \textbf{Ours (Full Pipeline)}: The complete three-phase pipeline (Phase~1 data synthesis + Phase~2 SFT + Phase~3 GRPO).
\end{itemize}

\subsection{Training Details}
We fine-tune Qwen3-8B using a two-stage pipeline: Supervised Fine-Tuning (SFT) followed by GRPO reinforcement learning. Both stages utilize 4-bit quantization and LoRA adapters ($r=16, \alpha=32$) on all linear layers via the Unsloth-TRL framework. SFT is conducted for 4 epochs with a 1e-4 learning rate and a 16,384 token context window. Subsequently, we perform 1 epoch of GRPO using curated data from Section~\ref{data-curation}, employing 8 generations per prompt ($G=8$) and a 5e-5 learning rate. Both phases use the AdamW optimizer, a cosine schedule, and bf16 mixed precision, with the RL phase capping prompt and completion lengths at 4,096 tokens each.
\section{Results}
\label{sec:results}

\subsection{Main Results}
\label{sec:results:main}

\begin{table}[t]
  \caption{Pass@1 (\%) on MultiPL-E and Ag-LCB. $\Delta$ shows improvement over the base model.}
  \label{tab:main_results}
  \centering
  \begin{tabular}{l | r r | r r}
    \toprule
    \multirow{2}{*}{\textbf{Model}} & \multicolumn{2}{c|}{\textbf{MultiPL-E}} & \multicolumn{2}{c}{\textbf{Ag-LCB}} \\
    & Julia & Bal. & Julia & Bal. \\
    \midrule
    Qwen3-8B (Base)         & 44.0 & 4.4 & 9 & 2.9 \\
    \quad + SFT Only        & 57.9 & \textbf{56.0} & \ 18.8 & 9.9 \\
    \midrule
    Ours (Full Pipeline)    & \textbf{68.6} & 49.7 & \textbf{39.2} & \textbf{25.0} \\
    \quad $\Delta$ vs.\ Base & +24.6 & +45.3 & +30.2 & +22.1 \\
    \bottomrule
  \end{tabular}
\end{table}

Table~\ref{tab:main_results} summarizes the pass@1 results for Julia and Ballerina across both benchmarks. For each language, we report the base model, the SFT-only variant, and the full three-phase pipeline. Since Agnostics~\cite{boruch-gruszecki2026agnostics} already provides a thorough evaluation of RL-only training across multiple configurations, we include the RL-only variant solely as an ablation under our specific setup (Section~\ref{sec:results:ablation}). The key finding is that each successive phase yields additive gains: SFT alone substantially improves over the base model by embedding syntactic priors, and GRPO further boosts performance by optimizing functional correctness. We discuss each language in detail below.

\subsubsection{Julia.}
Compared to vanilla Qwan mode baseline, our pipeline improves Julia pass@1 from 28.3\% to 68.6\% (+40.3 points) on MultiPL-E and from 9\% to 39.2\% (+30 points) on Ag-LCB. Table~\ref{tab:related_work_comparison} places these results in context. Most importantly, we surpass the previous state of the art (Agnostics~\cite{boruch-gruszecki2026agnostics}) by 7.6 percentage points on MultiPL-E and 14.2 percentage points on Ag-LCB, despite using a single base model rather than separate task-specific variants.

\begin{table}[t]
  \caption{Comparison with prior work on Julia code generation (pass@1 \%). Results for Giagnorio et al.\ and Agnostics are taken from the respective papers.}
  \label{tab:related_work_comparison}
  \centering
  \begin{tabular}{l l r r}
    \toprule
    \rotatebox{0}{\textbf{Method}} & \rotatebox{0}{\textbf{Base Model}} & \rotatebox{90}{\textbf{MultiPL-E}} & \rotatebox{90}{\textbf{Ag-LCB}}\\
    \midrule
    \multirow{2}{*}{Giagnorio et al.~\cite{giagnorio2025enhancing}} & DeepSeek Coder & 41.2  & -\\
    & CodeLlama-7B & 28.4 & -\\
    \midrule
    \multirow{2}{*}{Agnostics~\cite{boruch-gruszecki2026agnostics}} & Qwen3-4B-MBPP & 62 & 15\\
    & Qwen3-8B-CF & 61 & 25\\
    \midrule
    Ours (SFT Only) & Qwen3-8B & 57.9 & 18.8\\
    Ours (Full Pipeline) & Qwen3-8B & \textbf{68.6} & \textbf{39.2}\\
    \bottomrule
  \end{tabular}
\end{table}

\subsubsection{Ballerina.}

For the extreme-LRPL case, SFT increased ballerina scores 4.4\% to 56\% and LCB scores by 2.9\% to 9.9\%. GRPO training using a curated dataset increased ballerina scores for LCB from 9.9\% to 25\%, but reduced the MultiPL-E results to 49.7\%. We believe this is overall positive given that LCB is a much harder benchmark 

Applying the pipeline to Ballerina required minimal language-specific modifications: (1)~a working Ballerina compiler and runtime for sandbox execution, (2)~translation of seed problem descriptions to specify Ballerina-specific IO formats where necessary, and (3)~no changes to the GRPO configuration, as IO tests are inherently language-agnostic. The fact that our pipeline produces functional code generation capability from near-zero provides strong evidence that, given a compiler and language-agnostic problems, any language can be bootstrapped.

\subsubsection{Data and Cost Efficiency.}
These results are achieved with fewer than 1{,}000 SFT samples and approximately 500 GRPO training problems, a $3\times$ reduction in training data compared to Agnostics, which uses ${\sim}$5{,}000 GRPO samples. The cost advantage is equally striking: Agnostics requires \$320.3 ($13\text{h} \times 8\,\text{GPUs} \times \$3.08/\text{GPU-h}$), whereas our total cost is \$54.02, comprising fine-tuning ($45\text{h} \times \$0.42/\text{GPU-h}$), OpenAI GPT OSS-120B API calls ($5\text{M input} \times \$0.25/\text{1M}$ $+\; 3\text{M output} \times \$0.60/\text{1M}$), and Sonnet API calls ($4.1\text{M input} \times \$3/\text{1M}$ $+\; 1.3\text{M output} \times \$15/\text{1M}$). Our pipeline thus costs only $\frac{1}{6}$ of Agnostics. The key enablers are left-shifting inference-time compute to offline data synthesis and training with a compact, difficulty-filtered RL set. Furthermore, by left-shifting, rather than spending GPU hours on large-scale RL exploration, we invest a one-time API cost to curate high-quality SFT data.

\subsection{Error Category Analysis}
\label{sec:results:errors}

To understand \emph{why} each pipeline phase improves performance, we categorize errors from the base model, SFT-only model, and full pipeline on MultiPL-E problems for Julia.

\begin{table}[t]
  \caption{Error distribution (\%) on MultiPL-E across pipeline phases.}
  \label{tab:error_analysis}
  \centering
  \begin{tabular}{l r r r}
    \toprule
    \textbf{Error Category} & \textbf{Base} & \textbf{+SFT} & \textbf{Full} \\
    \midrule
    \multicolumn{4}{l}{\textit{Julia}} \\
    \quad Syntax / Compilation         & 45.9 & 0.6 & 0.6 \\
    \quad Runtime (type, bounds, etc.) & 7.5 & 13.8 & 9.4 \\
    \quad Wrong Output                 & 2.5 & 27.7 & 26.4 \\
    \quad Timeout                      & 0.0 & 0.0 & 0.0 \\
    \quad Correct                      & 44.0 & 57.9 & 63.5 \\
    \bottomrule
  \end{tabular}
\end{table}

Table~\ref{tab:error_analysis} reveals a clear progression. The base model's failures are dominated by syntax and compilation errors, the model frequently produces Python-influenced constructs that fail to compile in Julia. After SFT, syntax errors drop substantially, but runtime errors and wrong outputs increase, indicating that the model now produces valid Julia code that compiles but does not yet implement correct algorithms. The full pipeline further reduces wrong-output errors, as the GRPO phase optimizes for functional consistency within the space.

\section{Ablation Studies}
\label{sec:results:ablation}

As ablations to difficulty-based data curation, we performed two experiments with Julia. For that purpose, in addition to the curated dataset, we created \textit{a random difficulty dataset} by sampling 500 problems from the full ELO range (spanning ELO 800--2200). This acts as a control dataset as opposed to our curated dataset. 

As the first experiment, we started with the vanilla Qwan model (no-SFT) and then fine-tuned it with GRPO using the random difficulty dataset. 
As the second experiment, we started with the Qwan model already fine-tuned with SFT, and further fine-tuned it with GRPO using the random difficulty dataset. 

In the following discussion, we compare our default pipeline (Qwan model + SFT + GRPO with curated dataset).

\subsection{Default pipeline vs. experiment 1 (qwan + GRPO with random dataset)}

\begin{figure*}[t]
  \centering
  \includegraphics[width=\textwidth]{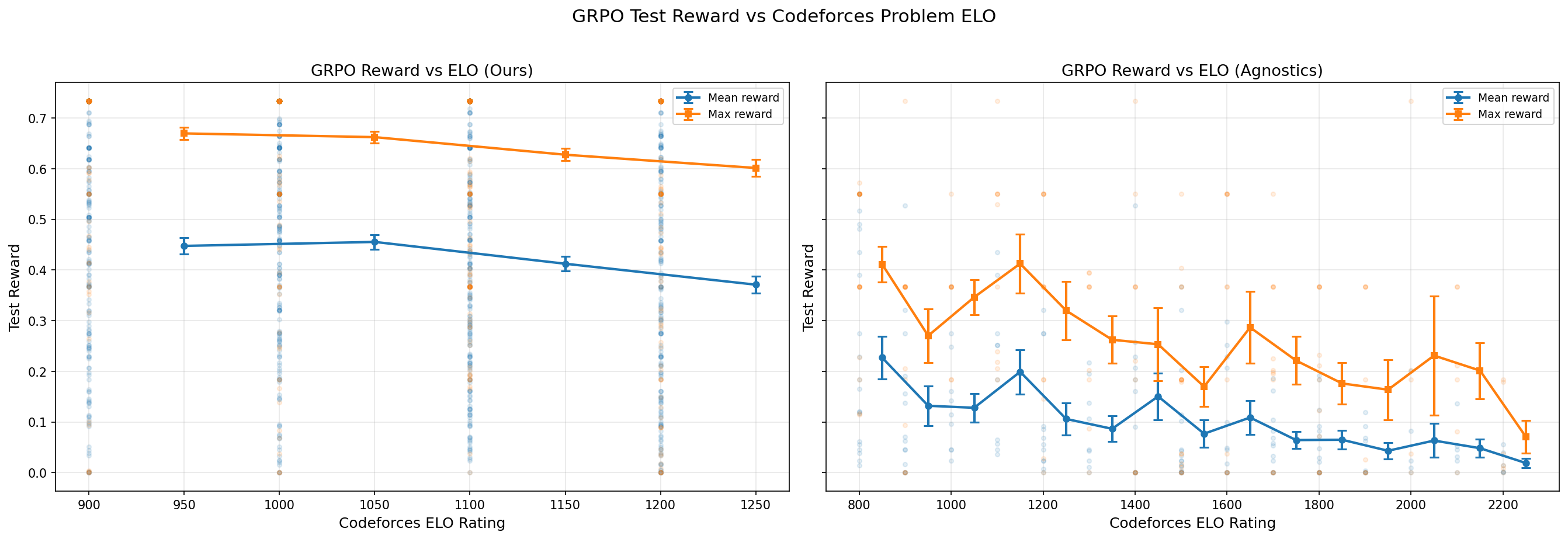}
  \caption{Comparison of GRPO reward vs.\ Codeforces problem ELO between our pipeline (\textbf{Left}) and the Agnostics pipeline~\cite{boruch-gruszecki2026agnostics} (\textbf{Right}). Our SFT-initialized policy achieves on average higher mean reward and maintains stable advantage signals across the training ELO range, indicating consistent and stable RL training. In contrast, the Agnostics pipeline exhibits degrading rewards and growing variance at higher ELO ratings.}
  \Description{...}
  \label{fig:reward_vs_elo}
\end{figure*}

The Figure~\ref{fig:reward_vs_elo} plots compares GRPO test reward against difficulty (ELO rating). The left-hand side shows our pipeline (Qwan + SFT + GRPO with curated dataset) produces remarkably stable reward signals: mean between 0.4 and 0.5 across the training ELO range (800- 1250), and max reward stays near 1.0, indicating that at least one completion per GRPO group fully solves each problem. This confirms our default pipeline provides effective rewards.

In contrast, the right-hand side of the Figure showing data from experiment 1 (no SFT and GRPO random dataset) exhibits substantially higher variance in both mean and max reward, particularly as problem difficulty increases beyond ELO~1400. The growing error bars reflect the instability of RL exploration when the policy must simultaneously discover language syntax and algorithmic logic. 

\subsection{Default pipeline vs. experiment 2 (qwan + SFT + GRPO with random dataset)}
Table~\ref{tab:selection} compares the Pass@1 scores between SFT mode trained with GRPO using curated vs random datasets. Random baseline yields 52.2\% pass@1, 16.4 percentage points below the difficulty-curated variant. These results confirm that for low-resource languages, curating RL training data by difficulty amplifies the effectiveness of GRPO by concentrating training signal on problems at the frontier of model capability.

\begin{figure}[h]
  \centering
  \includegraphics[width=\textwidth/2]{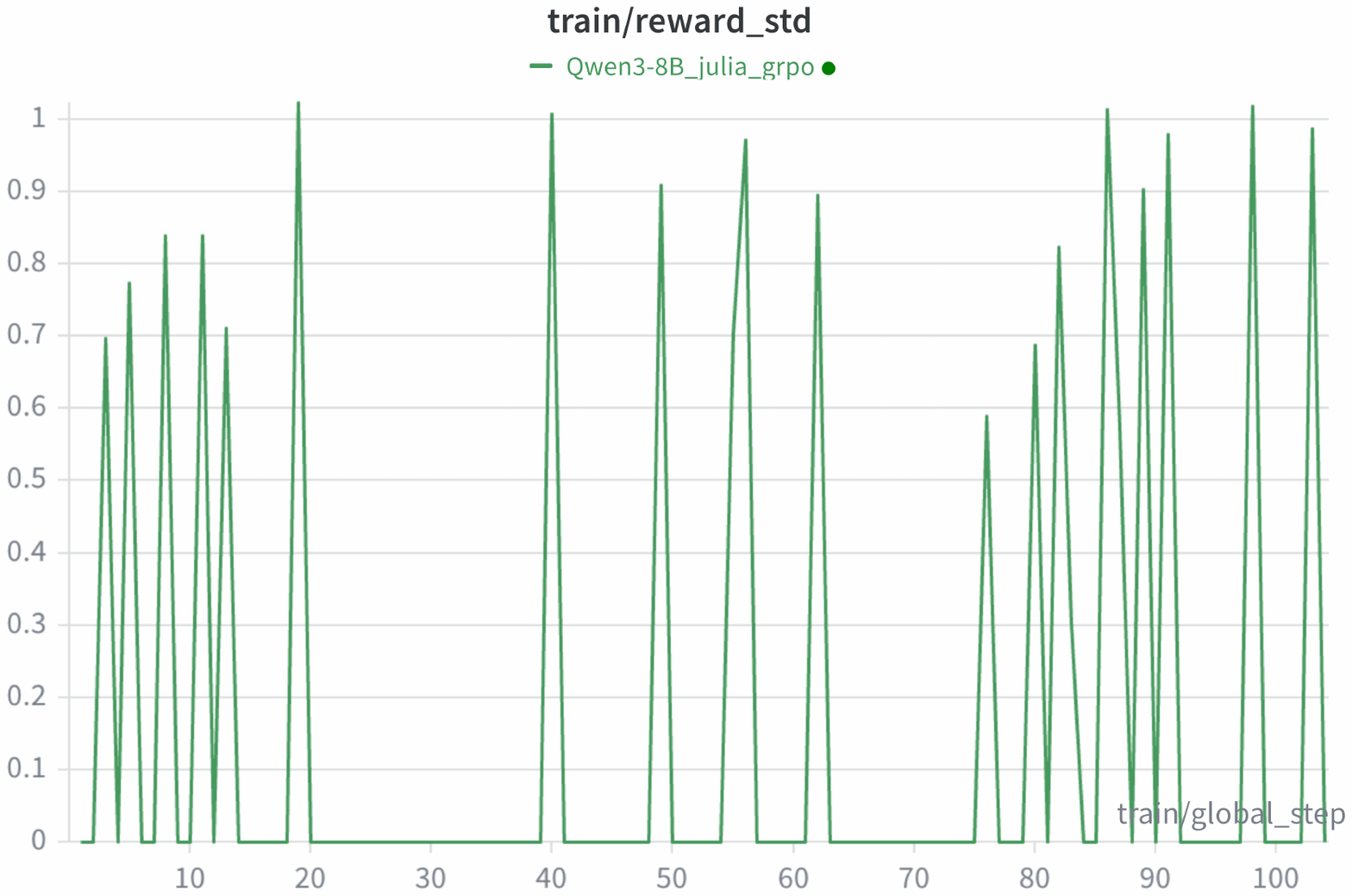}
    \caption{Reward standard deviation across training steps after zero-advantage masking when we loose the `deliberate practice'}
    \Description{...}

  \label{fig:reward_std}
\end{figure}


\begin{table}[t]
  \caption{Comparison of GRPO with problem selected randomly and ELO based selection}
  \label{tab:selection}
  \centering
  \begin{tabular}{ l r }
    \toprule
    \textbf{Method}  & \textbf{pass@1}\\
    \midrule
    
    Random Dataset  & 52.2\\
    Curated Dataset (ELO <1250)  & 68.6\\
    \bottomrule
  \end{tabular}
\end{table}


\section{Discussion}
\label{sec:discussion}

\subsection{Zero-advantage Masking}
Figure~\ref{fig:reward_std} illustrates this phenomenon empirically. After zero-advantage masking, the vast majority of training samples are discarded, visible as long stretches where $\sigma_G = 0$. This occurs when all completions within a group either uniformly succeed or uniformly fail, collapsing the advantage to zero. The isolated spikes where $\sigma_G > 0$ represent the small fraction of samples that actually produce a meaningful gradient signal, problems at the edge of the model's current capability. This sparse pattern highlights the inefficiency of training without difficulty-based curation: most computational effort is spent on samples that contribute nothing to learning.

\subsection{New Languages Require the SFT Phase}
The Ballerina experiment validates that the SFT phase is essential for severely low-resource programming language (S-LRPL) adoption. Without syntactic priors, the RL exploration space is too unconstrained: the model spends its compute budget rediscovering basic syntax rather than learning algorithmic patterns. This suggests that for any new LRPL, investing in Phase~1 data synthesis and Phase~2 SFT is a prerequisite before RL can be effective. The pipeline thus provides a practical recipe: given a compiler for language $L$ and a set of language-agnostic algorithmic problems, one can bootstrap code generation capability for $L$ through offline data synthesis followed by SFT and GRPO.

\subsection{Right Level of Difficulty Amplifies Pipeline Benefits}
Our difficulty-based curation results echo the principle of \emph{deliberate practice}~\cite{Ericsson2008DeliberatePA}: learning is most effective when training is concentrated at the boundary of current competence. The mechanism is intuitive within the GRPO framework. Given a problem, GRPO generates multiple candidate solutions and updates the policy based on the relative advantage between them. If a problem is too easy, all candidates succeed and the advantage collapses to zero; if too hard, all candidates fail with the same outcome. Only when problems lie at the edge of the model's capability, where some candidates succeed and others fail, does the advantage carry a meaningful gradient signal.

This follows directly from the GRPO advantage (Eq.~\ref{eq:advantage}). With binary pass/fail rewards $r_i \in \{0,1\}$:
\begin{itemize}
  \item \textbf{Too easy} ($r_i = 1\;\forall\, i$): $\mu_G = 1,\; \sigma_G = 0 \implies \hat{A}_i = 0$.
  \item \textbf{Too hard} ($r_i = 0\;\forall\, i$): $\mu_G = 0,\; \sigma_G = 0 \implies \hat{A}_i = 0$.
  \item \textbf{Edge of capability} (mixed pass/fail): $\sigma_G > 0$, so correct solutions receive $\hat{A}_i > 0$ and incorrect ones $\hat{A}_i < 0$, producing a meaningful gradient signal.
\end{itemize}
This provides a mathematical justification for difficulty-based data curation: selecting problems near the model's decision boundary maximizes the effective gradient from each GRPO update. This principle extends beyond LRPLs; Zhang et al.~\cite{Zhang2025BridgingVA} independently demonstrate that dynamically alternating between SFT and RL enables automatic weakness identification and targeted resource allocation, further corroborating the value of difficulty-aware training.

\subsection{Compute Allocation: Offline vs.\ Online}

Our results contribute to the broader debate about where to invest compute for code generation. The field currently allocates substantial resources to inference-time scaling, multiple generation rounds, self-correction loops, and agent-based frameworks~\cite{huang2023agentcoder, codesim2025}. Our pipeline demonstrates that the same iterative refinement mechanism yields strictly greater returns when deployed offline as a data synthesis engine. The key difference is amortization: offline synthesis invests compute once to produce training data, while inference-time scaling imposes a per-query cost. To illustrate, consider an IDE code-completion plugin serving 1{,}000 Julia queries per day. An inference-time scaling approach requiring five rounds of refinement per query would consume $5\times$ the compute of a single forward pass on every query, accumulating substantial cost over weeks. In contrast, our pipeline pays a fixed one-time cost of approximately \$54 for data synthesis and fine-tuning, after which every query is served at single-pass inference cost. The crossover point is reached within days of deployment, and the cost advantage compounds indefinitely. Moreover, the offline synthesis cost is dominated by API calls for the initial data generation, a component whose price continues to decline as model serving becomes more efficient, making the approach increasingly economical over time.

\subsection{Practical Deployment Considerations}
At 8B parameters, the resulting model can run on a single consumer GPU, making it suitable for on-premise deployment where organizations cannot send proprietary code to external APIs. Combined with the pipeline's language-agnostic design, this enables organizations to bootstrap code generation for internal or niche programming languages without relying on cloud-based LLM services.

\section{Conclusion and Future work}
\label{sec:conclusion}

We presented a three-phase pipeline for adapting Small Language Models to Low-Resource Programming Languages that reframes inference-time scaling as offline data synthesis. By decoupling syntax acquisition (via compiler-feedback SFT) from algorithmic reasoning (via GRPO with IO tests), our pipeline resolves the trilemma of data scarcity, inference cost, and sparse RL rewards. Furthermore, by curating the data for GRPO use, we amplify and stabilize the reward signal. Applied to Qwen3-8B, it exceeded state-of-the-art results on MultiPL-E (+7.6 pass@1) and Ag-LCB (+14.2 pass@1) for Julia. Furthermore, the pipeline generalizes to Ballerina, a language with near-zero pretraining representation, achieving Pass@1 56\% with MultiPL-E and 25\% with LCB. 

Our ablation studies confirm that iterative compiler-feedback data synthesis is essential for high-quality SFT, as it stabilizes the reward, eliminates the subtle errors present in naively generated synthetic data, and yields a cleaner training signal.


Future research will focus on scaling the pipeline across diverse model architectures and expanding its support for a wider range of low-resource programming languages. We plan to investigate the synergy between continued pretraining and SFT, develop automated metrics for code idiomaticity, and integrate offline data synthesis with inference-time refinement to further boost performance.

\section*{Data Availability}

The source code and datasets supporting the findings of
this study are available at the following Zenodo repository \href{https://zenodo.org/records/19349053?preview=1&token=eyJhbGciOiJIUzUxMiJ9.eyJpZCI6ImQwNTIyYzU1LTMzMTItNGFhMC1iODdiLTRmNmM5NGNlNzc1NCIsImRhdGEiOnt9LCJyYW5kb20iOiJiNGYzNTYwZDllMTM5Zjc1NzQzNGYyMzU1ZDgxZDM2OSJ9.m-NBg_M6AZjJ2c3fRZiI0cdeJwWeTyXH8qDETYcfXSbbbMOtU537NVTT9p3ge_9yhECe_F5XmlduEvSjfTN38w}{[Click Here]}.

\newpage

\printbibliography

\end{document}